\documentclass[10pt, a4paper]{article}
\usepackage{lrec}
\usepackage{graphicx}
\usepackage{tabularx}
\usepackage{soul}
\usepackage{xcolor}
\usepackage{url}
\usepackage{epstopdf}
\usepackage[utf8]{inputenc}

\usepackage{hyperref}
\usepackage{xstring}

\title{A Swiss German Dictionary: \\ Variation in Speech and Writing}
\name{Larissa Schmidt$^{\ast}$, Lucy Linder$^{\dagger}$, Sandra Djambazovska$^{\dagger}$$^{\dagger}$, \\ {\bf \large Alexandros Lazaridis$^{\dagger}$$^{\dagger}$,  Tanja Samardžić$^{\ast}$, Claudiu Musat$^{\dagger}$$^{\dagger}$ }}

\address{$^{\ast}$ University of Zurich, URPP Language and Space \\
$^{\dagger}$ University of Fribourg \\
$^{\dagger}$$^{\dagger}$ Data, Analytics and AI (DNA) - Swisscom AG   \\ $^{\ast}$ $ \{ $larissa.schmidt, tanja.samardzic$ \} $@uzh.ch,  $^{\dagger}$ lucy.linder@hefr.ch, \\ $^{\dagger}$$^{\dagger}$ sandra.djambazovska@gmail.com, $ \{ $alexandros.lazaridis, claudiu.musat$ \} $@swisscom.com }

\abstract{
We introduce a dictionary containing forms of common words in various Swiss German dialects normalized into High German. As Swiss German is, for now, a predominantly spoken language, there is a significant variation in the written forms, even between speakers of the same dialect. To alleviate the uncertainty associated with this diversity, we complement the pairs of Swiss German - High German words with the Swiss German phonetic transcriptions (SAMPA). This dictionary becomes thus the first resource to combine large-scale spontaneous translation with phonetic transcriptions. Moreover, we control for the regional distribution and insure the equal representation of the major Swiss dialects.
The coupling of the phonetic and written Swiss German forms is powerful. We show that they are sufficient to train a Transformer-based phoneme to grapheme model that generates credible novel Swiss German writings. In addition, we show that the inverse mapping - from graphemes to phonemes - can be modeled with a transformer trained with the novel dictionary. This generation of pronunciations for previously unknown words is key in training extensible automated speech recognition (ASR) systems, which are key beneficiaries of this dictionary.
\\ \newline }

\begin{document}

\maketitleabstract

\section{Introduction}\label{sec:intro}

\textit{Swiss German} refers to any of the German varieties that are spoken in about two thirds of Switzerland \cite{archimob_16}. Besides at least one of those dialectal varieties, Swiss German people also master standard (or 'High') German which is taught in school as the official language of communication. \\
Swiss German is varies strongly. Many differences exist in the dialectal continuum of the German speaking part of Switzerland. Besides pronunciation, it also varies a lot in writing. Standard German used to be the exclusive language for writing in Switzerland. Writing in Swiss German has only come up rather recently (notably in text messaging). Because of this, there are no orthographic conventions for Swiss German varieties. Even people speaking the same dialect can, and often do, write phonetically identical words differently. \\
In this paper, we present a dictionary of written standard German words paired with their pronunciation in Swiss German words. Additionally Swiss German spontaneous writings, i.e. writings as they may be used in text messages by native speakers, are paired with Swiss German pronunciations. \\
The primary motivation for building this dictionary is rendering Swiss German accessible for technologies such as Automatic Speech Recognition (ASR). \\
This is the first publicly described Swiss German dictionary shared for research purposes. Furthermore, this is the first dictionary that combines pronunciations of Swiss German with spontaneous writings.

\section{Related Work}\label{sec:relatedwork}
%\subsection{Corpora for Swiss German }\label{subsec:existing_corpora}

\begin{figure}[!h]
\begin{center}
%\fbox{\parbox{6cm}{
%This is a figure with a caption.}}
\includegraphics[scale= 0.3]{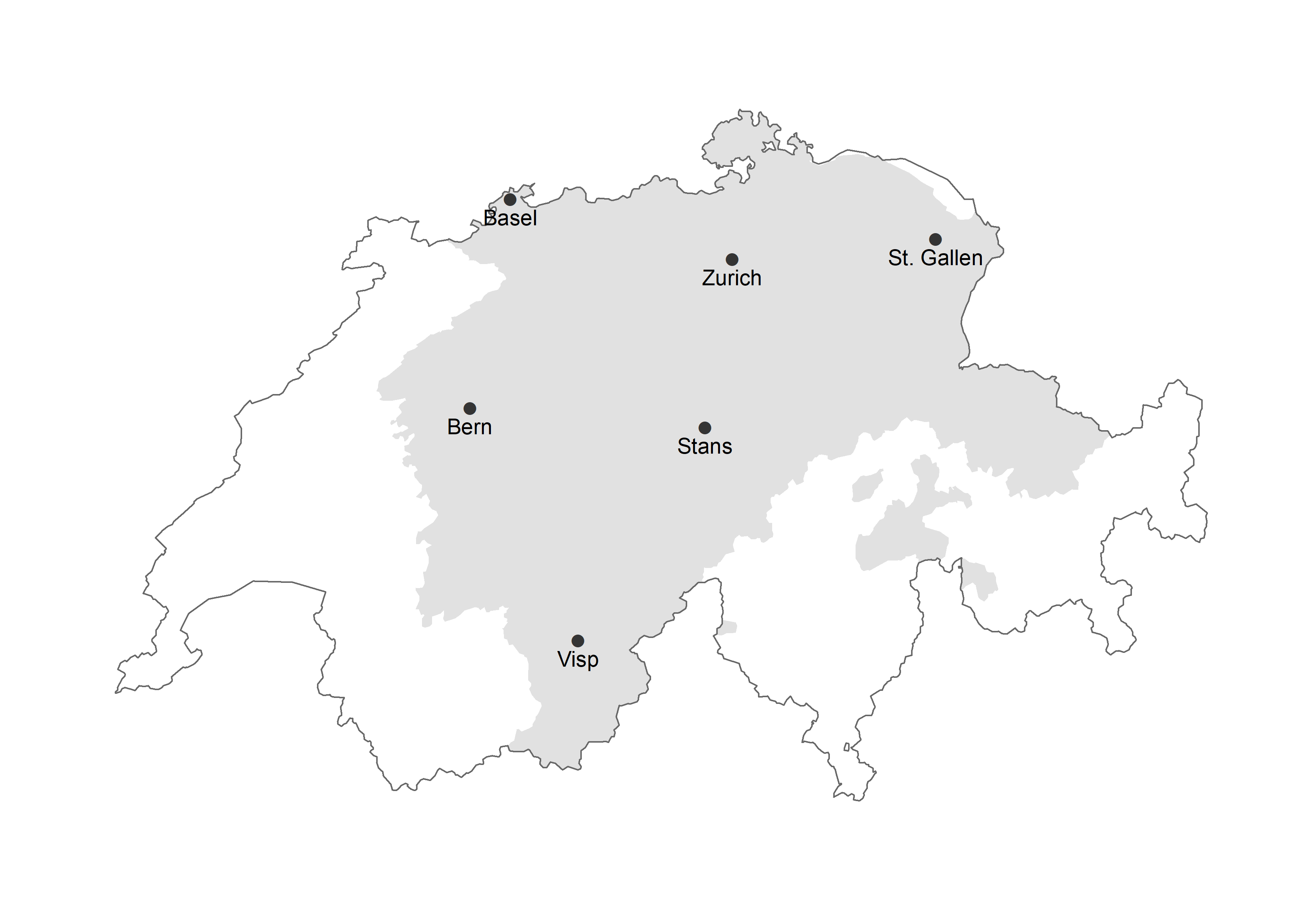} 
\caption{Six variants of Swiss German chosen for our dictionary. Map by Yves Scherrer and Larissa Schmidt.}

\label{fig.1}
\end{center}
\end{figure}
This dictionary complements previously developed resources for Swiss German, which share some common information. Spontaneous noisy writing has already been recorded in text corpora \cite{noah,sms_corpus,wus_corpus}, some of which are also normalized. These resources contain relatively large lexicons of words used in context, but they do not contain any information about pronunciation. The features of speech are represented in other resources, such as \cite{sds,dialaekt_app,voiceapp}, which, on the other hand, contain relatively small lexicons (small set of words known to vary across dialects). The ArchiMob corpus does contain a large lexicon of speech and writing (Dieth transcription), but the spoken part is available in audio sources only, without phonetic transcription. 

This dictionary is the first resource to combine all the relevant information together. A relatively large lexicon has been constructed in which phonetic transcriptions (in the SAMPA alphabet) are mapped to various spontaneous writings controlling for the regional distribution. Some of the representations in this dictionary are produced manually, while others are added using automatic processing.

%\subsection{Swiss German seq2seq models}\label{subsec:SwissGerman_seq2seq}
%TS first draft
Automatic word-level conversion between various writings in Swiss German has been addressed in several projects, mostly for the purpose of writing normalization \cite{idiotikon,sms_corpus,wus_14,lusettietal_18,ruzsicsetal_19,archimob_15,archimob_16,archimob_19}. The task of normalization consist of mapping multiple variants of a single lexical item into a single writing usually identical to standard German (an example would be the Swiss German words \textit{aarbet} and \textit{arbäit} which both map to standard German \textit{arbeit} ('work')). Early data sets were processed manually (SMS). This was followed by an implementation of character-level statistical machine translation models \cite{Samardzicetal2015,scherrer16-automatic} and, more recently, with neural sequence-to-sequence technology. The solution by \newcite{lusettietal_18} employes soft-attention encoder-decoder recurrent networks enhanced with synchronous multilevel decoding. \newcite{ruzsicsetal_19} develop these models further to integrate linguistic (PoS) features. 

A slightly different task of translating between standard German and Swiss dialects was first addressed with finite state technology \cite{scherrer_12}. More recently, \newcite{honnet-etal17} test convolutional neural networks on several data sets.

We continue the work on using neural networks for modeling word-level conversion. Unlike previous work, which dealt with written forms only, we train models for mapping phonetic representations to various possible writings. The proposed solution relies on the latest framework for sequence-to-sequence tasks --- transformer networks \cite{vaswani2017attention}.

%The following is now the section on data
\section{Dictionary Content and access}\label{sec:data}
\begin{table*}[ht]
\begin{center}
\begin{tabular}{llllllll}
%the examples show
      %\hline
      \textbf{standard} & \textbf{S/W} & \textbf{Zurich}& \textbf{St. Gallen} &\textbf{Basel}& \textbf{Bern} & \textbf{Visp} & \textbf{Stans} \\
      \hline%\hline
      %rauchen ('smoke')  & S & r aU x @ &  R aU x E &  R Eu kh @  &  r oU x @ &  r 9I kx U &  r AI kx @ \\
      %that Swiss German has a lot of different diphthongs – (problem for processing, eased by reduction of phoneset) important in this section?
       %& W & rauche & rauche & rauke & rouche & röiku & raike\\
      %\hline
      liebe ('love') & S & l i @ b @  & l i @ b E &  l I @ b I  &  l i @ b i &  l I$ \{ $ b I & l IE b I \\ %diphthongs not consistent – problem?
      & W & liebi & liebe & liebe & liebe & liebu & liebe\\
      %the morphological ambiguity of a lot of standard german words
      \hline
       frage ('question') & S & f r 2: g @  & f R O: g &  f R O: g  &  f r a: g &  f r a: k & f r A: k \\
       & W & frag & froog & froog & fraag & freegu & fraag\\
       %invent GSW myself
      \hline
       %bank ('bank')& S & b a N kx  & b a N kx &  b a n k  &  b a N kx &  b oI C & b A N kx \\
       %& W & gsw zh & gsw sg & gsw bs & gsw be & gsw vs & gsw nw\\
       %\hline
      lecker ('tasty') & S & f aI n  & f aI n &  l E kh @ R  &  f eI n & f $ \{$I n  & f $ \{$I n  \\
       & W &fein& fein & lecker & fein & lecker & fein\\
      \hline
       rasch ('swiftly') & S & r a S  & R a S &  R a S  &  r a S &  k S v I n t & t I f I k \\
        & W & schnäll & rasch & rasch & rasch & gschnäll & tifig\\
      \hline
      % frühstück ('breakfast') & S & ts m o r g @	& ts m O R g E	& ts m O R g @	& ts m o r g @	& f r Y$ \{$ S t U kx&	kx A l a ts @ \\
       %& W & gsw zh & gsw sg & gsw bs & gsw be & gsw vs & gsw nw\\
      %\hline
       %erhalten ('receive') & S & E r h a l t @ &	E R h a l t E &	 e R h a l t @	& e r h a w t @ &	 f E r C o &	 E r h AU t @ \\
      % & W & becho & gsw sg & gsw bs & gsw be & vercho & gsw nw\\
      %\hline
      %gsws above are spontaneous from me and raphi
       ging ('went') & S & b i n k a N @ &	b i n k a N @ &	b I k a N @ &	b @ k a N @ &	I S k a N U &	I S k A N @ \\
       & W & bi gange & bin gange & bi gange & bi gange & bi gangu & isch gange\\
      %\hline
       %gossau (toponym) & S &	g o: s aU &	g o: s aU &	g O ss aU &	g o ss oU &	k o ss aU &	g o ss AU \\
       %& W & gsw zh & gsw sg & gsw bs & gsw be & gsw vs & gsw nw\\
      \hline

\end{tabular}
\caption{Dictionary entry of five standard German words mapped with their spoken (=S) Swiss German representation (in SAMPA) toghether with a Swiss German spontaneous writing (=W) in the six dialects of Zurich, St. Gallen, Basel, Bern, Visp, and Stans}
\label{tab:dictionary_entry}
\end{center}
\end{table*}

%standard German written <> Swiss German phonetic word
%number of phonetic words
%Swiss German phonetic <> Swiss German writing
We pair 11'248 standard German written words with their phonetical representations in six different Swiss dialects: Zürich, St. Gallen, Basel, Bern, Visp, and Stans (Figure \ref{fig.1}). The phonetic words were written in a modified version of the Speech Assessment Methods Phonetic Alphabet (SAMPA). The Swiss German phonetic words are also paired with Swiss German writings in the latin alphabet. (From here onwards, a phonetic representation of a Swiss German word will be called \textit{a SAMPA} and a written Swiss German word will be called \textit{a GSW}.)

%big set, small set – reasons for both
This dictionary comes in two versions as we used two differently sized sets of SAMPA characters. Our extended set including 137 phones allows for a detailed and adequate representation of the diverse pronunciation in Switzerland. The smaller set of 59 phones is easier to compute. The phone reduction was mainly done by splitting up combined SAMPA-characters such as diphthongs. \textbf{UI} s t r $ \{$ tt @ and \textbf{U I} s t r $ \{$ t t @ for example are both representations of the Stans pronunciation of the standard German word \textit{austreten} ('step out'). The latter representation belongs to the dictionary based on the smaller phoneset. Table \ref{tab:dictionary_entry} shows an example of five dictionary entries based on the bigger phoneset. 
%zahlen nochmals checken

%How many writings manually
For a subset of 9000 of 11'248 standard German words, we have manually annotated GSWs for Visp (9000) and for Zurich (2 x 9000, done by two different annotators). For a subsubset of 600 of those standard German words we have manually annotated GSWs for the four other dialects of St. Gallen, Basel, Bern, and Stans. The remaining writing variants are generated using automatic methods described below.
 
 The dictionary is freely available for research purposes under the creative commons share-alike non-commercial licence\footnote{\url{https://creativecommons.org/licenses/by-nc-sa/4.0/}} via this website \url{http://tiny.uzh.ch/11X}.

%\subsection{Access to dictionary}\label{subsec:access}

\section{Construction of the dictionary}\label{sec:methods}
\begin{table*}[ht]
\begin{center}
\begin{tabular}{llll}
\textbf{english gloss} & \textbf{correct version} & \textbf{wrong version} & \textbf{error} \\
\hline
plant & pflanze[1] & pflanz\textit{l}e[0] &  at least one additional letter \\
Peter & peter[1] & pter[0] & at least one missing letter \\
groom & rossknächt[1] & rosskn\textit{a}cht[0] & at least one changed letter \\
%about 'changed': 
%% this includes: all cases where the letter stands in no relation to the sound (such as given in the above example). but it also includes some cases where there is a relation between grapheme and phoneme. For example, the letter 'c' stands in in a relation to the sound (SAMPA) s, as it is often used for representing that sound. Yet for the word 'setzt', we counted 'cetzt' as wrong, because it is too far off from german writing. Same goes for 
    %\begin{itemize}
    %         \item  Grapheme \textit{ch} 
    %         %instead of \textit{ck}, 
    %         for sound (in SAMPA) \textit{kx}, e.g. [0] \textit{wipchingu} ('Wipkingen'), or [0] \textit{wichel} ('poultice'). 
    %         \item Grapheme \textit{c} for sound (in SAMPA) \textit{s}, e.g. [0] \textit{cetzt} ('sits')
    %         \item Grapheme \textit{ch} for sound (in SAMPA) \textit{S} e.g. [0] \textit{chaffner} ('conductor')
    %     \end{itemize}{}
pension fund & pensionskasse[1] & pen\textit{ss}ionska\textit{s}e[0] & at least two 'minor' mistakes

\end{tabular}
\caption{Examples of evaluated GSWs. The 'correct version' is only one of many possible versions of GSWs, tagged '1' in our evaluation. The 'wrong version' was tagged '0' in our evaluation. The column 'error' shows the criteria we used for evaluating the GSWs as '0'.}
\label{tab:evaluation_criteria}
\end{center}
\end{table*}

In the following we present the steps of construction of our dictionary, also detailing how we chose the six dialects to represent Swiss German and how, starting with a list of standard German words, we retrieved the mapping SAMPAs and GSWs.  

\subsection{Discretising continuous variation}

To be able to represent Swiss German by only a few dialects which differ considerably it is necessary to discretize linguistic varieties. Because, as mentioned earlier, regional language variation in Switzerland is continuous. For this identification of different varieties we used a dialectometric analysis~\cite{scherrer_stoeckle16}. This analysis is based on lexical, phonological, morphological data of the German speaking areas of Switzerland~\cite{sds}. As we worked with word-lists and not sentences, we discounted syntactical influences on area boundaries that are also described in that analysis. %they also used sads data, did not refer to that as we did not use it, as mentioned. LS

We represent six differentiated linguistic varieties. We considered working with ten linguistic varieties because this number of areas was the 'best-cut'-analysis in the dialectometric analysis \cite[p.105]{scherrer_stoeckle16}. Yet, due to time restraints and considerable overlap between some of the linguistic varieties, we reduced this number to six. 
%The six varieties, suggested by Yves are Wallis and German speaking part of Grisons,the Bernese Oberland, the rest of Bern, Central Switzerland,Zurich and St. Gallen,Basel and surrounding area. – These are not in the paper, but from the Mail Yves sent me with additional computations
We also made some adjustements to the chosen varieties in order to correspond better to the perception of speakers and in favor of more densely populated areas.  
% We chose not to differentiate Bern and Berner Oberland by focusing exclusively on Bern. Instead, we chose to differentiate Zürich and St. Gallen. Furthermore, St. Gallen as a city is a more densely populated area than Berner Oberland with its many different dialects is, thus rendering St. Gallen a more numerically representative dialect variation.

% The six areas we chose to represent in the end are: Wallis and German speaking part of Grisons,the Bernese Oberland together the rest of Bern, Central Switzerland, Zurich, St. Gallen, Basel and surrounding area.

One way to represent the six individualized linguistic varieties would have been to annotate the dialectal centers, i.e. those places that have the average values of dialectal properties within the area where the variety is spoken. 
%%%%deleted%%%%%%%%%%%%%%%%%%%%%%%%%%%%%%%%%%%%
%  Yves Scherrer provided us with the list of places that had the average values of dialectal properties. For the 10-cluster analysis, these were: 
% \begin{itemize}
%     \item Meikirch (BE)
%     \item Illnau (ZH)
%     \item Römerswil (LU)
%     \item Alpnach (OW) 
%     \item Näfels (GL)
%     \item Reichenbach (BE) 
%     \item Kirchberg (SG)
%     \item Gelterkinden (BL)
%     \item Davos (GR)
%     \item Brig (VS)
% \end{itemize}
%%%%%%%%%%%%%%%%%%%%%%%%%%%%%%%%%%%%%%%%%%%%%%%%%
However, we chose to represent the linguistic varieties by the most convenient urban places. Those were the dialects of the Cities Zurich, St. Gallen, Basel, Bern, and Visp, and Stans.

\subsection{Manual annotation}\label{subsec:manual} %TS20191024 work -> annotation  

\subsubsection{SAMPAs}
For each standard German word in our dictionary we manually annotated its phonetic representation in the six chosen dialects. The information about the pronunciation of Swiss German words is partially available also from other sources but not fully accessible \cite{sds} \cite{idiotikon}.

To help us with pronunciation our annotators first used their knowledge as native speakers (for Zurich and Visp). Secondly, they consulted dialect specific grammars \cite{FleischerSchmid_2006} \cite{marti_1985} \cite{suter_1992} \cite{bohnenberger} \cite{Hug_weibel_2003} 
as well as dialect specific lexica 
\cite{osterwalder-braendle_2017} \cite{gasser_etal_2010}  \cite{niederberger_2007}. 
They also considered existing Swiss German dictionaries \cite{idiotikon} \cite{sds}, 
listened to recordings 
\cite{archimob_16}
and conferred with friends and acquaintances originating from the respective locations. 
 
 \subsubsection{GSWs}\label{subsubsec:gsw}
9000 GSWs for Visp German and 2 x 9000 GSWs for Zurich German were annotated by native speakers of the respective dialect. Our annotators created the GSWs while looking at standard German words and without looking at the corresponding SAMPAs for Visp and Zurich. Through this independence from SAMPAs we are able to avoid biases concerning the phonetics as well as the meaning of the word in generating GSWs.
 
At a later stage of our work, we added each 600 GSWs for the four dialects of St. Gallen, Basel, Bern, and Stans in order to improve our phoneme-to-grapheme(p2g) model (see next section). 
% The writings which the model generated in the first stage were not good – mainly because there were some special SAMPA characters in those dialects. 
For the manual annotation of these dialects we had no native speakers. Therefore, when writing the GSWs, our annotators relied on the corresponding SAMPAs of these dialects, which they had made an effort to create before. 

    % In order to account for the mentioned variety of everyday Swiss German writing, we aimed for more than one GSW per SAMPA. The heterogeneous writing style makes the SAMPA - GSW writing a one to many relation instead of the regular one to one that speakers of standard languages are accustomed to. To save time in generating the many GSWs, we opted for an automatic process. 
     
    %  We first tried to automatize the generation of GSWs with a rule-based program. Via SAMPAs together with phoneme-to-grapheme mappings we tried to obtain all possible GSWs. Yet, this yielded mostly impossible writings and also not all the writings we had already done manually.   
    
    % We then set up a phoneme-to-grapheme(p2g) model to generate the most likely spellings. The model was trained on the manually generated GSWs and the matching SAMPAs (see next section). 
    % The writings for the dialects of St. Gallen, Basel, Bern, and Stans which this model delivered in the first stage were not good – mainly because there were some special SAMPA characters in those dialects.
    
    % To improve the model, we added each 600 GSWs for the four dialects of St. Gallen, Basel, Bern, and Stans. For the manual annotation of these dialects we had no native speakers. Therefore, when writing the GSWs, our annotators relied on the corresponding SAMPAs of these dialects, which they had made an effort to create before. 

\subsection{Automatic annotation}\label{subsec:automatic}

In order to account for the mentioned variety of everyday Swiss German writing, we aimed for more than one GSW per SAMPA. The heterogeneous writing style makes the SAMPA$\,\to\,$GSW a one to many relation instead of the regular one to one that speakers of standard languages are accustomed to. To save time in generating the many GSWs, we opted for an automatic process. 
     
     We first tried to automatize the generation of GSWs with a rule-based program. Via SAMPAs together with phoneme-to-grapheme mappings we tried to obtain all possible GSWs. Yet, this yielded mostly impossible writings and also not all the writings we had already done manually. We then set up a phoneme-to-grapheme(p2g) model to generate the most likely spellings.

\subsubsection{Transformer-based Phoneme to Grapheme (p2g) }\label{subsubsec:modelp2g}

The process of generating written forms from a given SAMPA can be viewed as a sequence-to-sequence problem, where the input is a sequence of phonemes and the output is a sequence of graphemes.

We decided to use a Transformer-based model for the phoneme-to-grapheme (p2g) task. The reason for this is twofold. First, the Transformer has
shown great success in seq2seq tasks and it has outperformed LSTM and CNN-based models. Second, it is computationally more efficient than LSTM and CNN networks. 

The Transformer consists of an encoder and a decoder part. The encoder generates a contextual representation for each input SAMPA that is then fed into the decoder together with the previously decoded grapheme. They both have N identical layers. In the encoder, each layer has a multi-head self-attention layer and a position-wise fully-connected feed-forward layer. While in the decoder, in addition to these two layers, we also have an additional multi-headed attention layer that uses the output of the encoder \cite{vaswani2017attention}.

We are using a Pytorch implementation\footnote{\url{https://github.com/jadore801120/attention-is-all-you-need-pytorch}} of the Transformer. As a result of the small size of the dataset, we are using a smaller model with only 2 layers and 2 heads. The dimension of the key (d\_k) and value (d\_v) is 32, the dimension of the model (d\_model) and the word vectors (d\_word\_vec) is 50 and the hidden inner dimension (d\_inner\_hid) is 400. The model is trained for 55 epochs with a batch size of 64 and a dropout of 0.2. For decoding the output of the model, we are using beam search with beam size 10. We experimented with different beam sizes, but we saw that it does not have significant influence on the result.

The training set is made of 24'000 phonemes-to-graphemes pairs, which are the result of transcribing 8'000 High German words into two Zurich forms and one Visp form. Those transcriptions were made independently by three native speakers.
Due to the scarcity of data, we decided not to distinguish between dialects. Hence, a single model receives a sequence of SAMPA symbols and learns to generate a matching sequence of characters.

\subsubsection{Test set and evaluation}\label{subsubsec:evaluation}

Our team of Swiss German annotators evaluated a test-set of 1000 words. We aimed to exclude only very far-off forms (tagged '0'), such that they are very probably to be seen as false by Swiss German speakers. The accepted writings (tagged '1') might include some that seem off to the Swiss German reader.

 In order to consistently rate the output, the criteria shown in table \ref{tab:evaluation_criteria} were followed. A GSW was tagged '0' if there was at least one letter added, missing, or changed without comprehensible phonetic reason. GSWs were also tagged '0' if there were at least two mistakes that our annotators saw as minor. 'Minor mistakes' are substitutions of related sounds or spellings, added or omitted geminates, and changes in vowel length.

For each of the 1000 words in the test-set, five GSW-predictions in all six dialects were given to our annotators. For Visp and Zurich they tagged each 1000x5 GSW predictions with 1 or 0. For St. Gallen, Basel, Bern, and Stans, they evaluated 200x5. 

In Table \ref{tab:eval_01} we show the result from this evaluation. We count the number of correct GSWs (labeled as '1') among the top 5 candidates generated by the p2g model, where the first candidate is the most relevant, then the second one and so on. 
%In total we have 5000 words for Zurich and Wallis, 1000 from each candidate and 1000 words in the St. Gallen, Basel, Bern, and Nidwalden dialect, 200 from each candidate. 

The evaluation was done at a stage where our model was trained only on GSW for Zurich and Visp (see sec. \ref{subsubsec:gsw}). The amount of correct predictions are lower for the dialects of St. Gallen, Basel, Bern, and Stans, mainly because there were some special SAMPA characters we used for those dialects and the model did not have the correlating latin character strings. After the evaluation, we added each 600 GSWs for the four dialects of St. Gallen, Basel, Bern, and Stans to improve the model.

\begin{table*}[ht] 
\centering
\begin{tabular}{lrrrrrr}

%in Percent
      & \textbf{Visp} & \textbf{Zurich } & \textbf{Basel} & \textbf{St. Gallen} & \textbf{Bern} & \textbf{Stans}\\
      \hline%\hline
     	1st   &94.6 &87.3 &40.5     &54.5   &89.5   &58.5   \\
     	\hline
        2nd   &74.2 &57.3 &22.5     &35     &64.5   &42     \\
      \hline
      3rd     &61.4 &43.4 &22       &31.5   &42     &29     \\
      \hline
      4th     &55.8 &36.8 &13       &25.5   &36     &34     \\
      \hline
      5th     &48.4 &30.2 &16.5     &25     &29.5   &25     \\
      \hline\hline
      Total   &66.88 &51 &22.9      &34.3   &52.3   &37.7   \\
      \hline

    %   in numbers
    %   & \textbf{Visp} & \textbf{Zurich } & \textbf{Basel} & \textbf{St. Gallen} & \textbf{Bern} & \textbf{Stans}\\
    %   \hline%\hline
    %  	1st & 946 & 873 &81 & 109& 179& 117 \\
    %  	\hline
    %     2nd & 742 &573  & 45 & 70& 129& 84 \\
    %   \hline
    %   3rd & 614 & 434 & 44 & 63& 84& 58\\
    %   \hline
    %   4th & 558 & 368 & 26 &51 & 72& 68\\
    %   \hline
    %   5th &484  &302  & 33& 50& 59&50\\
    %   \hline\hline
    %   Total & 3344 & 2550 &229 & 343& 523& 377\\
    %   \hline

\end{tabular}
\caption{Percentages of correct GSWs among the top 5 candidates. For Zurich and Visp the total number of evaluated words was 5000, 1000 from each candidate. For St. Gallen, Basel, Bern, and Stans the total number of evaluated words was 1000, 200 from each candidate.}
\label{tab:eval_01}
\end{table*}

\subsubsection{Grapheme to Phoneme (g2p) and its benefits for ASR}\label{subsubsec:retraining}

Automatic speech recognition (ASR) systems are the main use cases for our dictionary. ASR systems convert spoken language into  
text. Today, they are widely used in different domains from customer and help centers to voice-controlled assistants and devices. The main resources needed for an ASR system are audio, transcriptions and a phonetic dictionary. The quality of the ASR system is highly dependant of the quality of the dictionary. With our resource we provide such a phonetic dictionary. 

To increase the benefits of our data for ASR systems, we also trained a grapheme-to-phoneme (g2p) model: Out-of-vocabulary words can be a problem for ASR system. For those out-of-vocabulary words we need a model that can generate pronunciations from a written form, in real time. This is why we train a grapheme-to-phoneme (g2p) model that generates a sequence of phonemes for a given word. We train the g2p model using our dictionary and compare its performance with a widely used joint-sequence g2p model, Sequitur \cite{Sequir}. For the g2p model we are using the same architecture as for the p2g model. The only difference is input and output vocabulary. The Sequitur and our model are using the dictionary with the same train (19'898 samples), test (2'412 samples) and validation (2'212 samples) split. Additionally, we also test their performance only on the items from the Zurich and Visp dialect, because most of the samples are from this two dialects. In Table \ref{tab:sequitur_vs_transformer} we show the result of the comparison of the two models. We compute the edit distance between the predicted and the true pronunciation and report the number of exact matches. In the first columns we have the result using the whole test set with all the dialects, and in the 2nd and 3rd columns we show the number of exact matches only on the samples from the test set that are from the Zurich and Visp dialect. For here we can clearly see that our model performs better than the Sequitur model. The reason why we have less matches in the Visp dialect compared to Zurich is because most of the our data is from the Zurich dialect. 
\begin{table*}[h] 
\centering
\begin{tabular}{lrrr}

      %\hline
      & \textbf{all (2412)} & \textbf{Zurich (1294)} & \textbf{Visp (825)} \\
      \hline%\hline
     	Transformer & 978 & 647 & 272\\
     	\hline
        Sequitur & 795 & 500 & 255\\
      \hline

\end{tabular}
\caption{Number of exact matches, Sequitur vs Transformer }
\label{tab:sequitur_vs_transformer}
\end{table*}

\section{Discussion}

%We came upon some limitations to our objectives. 

One of our objectives was to map phonetic words with their writings. There are some mismatches between SAMPA and GSWs in our dictionary, especially when the GSWs were done manually and independently from the SAMPA. Those mismatches occur where there is no straightforward correspondence of a standard German and Swiss German word. 

Two kinds of such a missing correspondence can be distinguished. First, there are ambiguous standard German words. And that is necessarily so, as our dictionary is based on a list of standard German words without sentential or any other context. 
%As we did not differentiate upper- and lower-case our dictionary is even more susceptible to ambiguity. 
An example for a (morphologically) ambiguous word is standard German \textit{liebe}. As we did not differentiate upper- and lower-case, it can both mean (a) 'I love'  or (b) 'the love'. As evident from table 1, \textit{liebe} (a) and \textit{liebi} (b) were mixed in our dictionary. The same is the case for standard German \textit{frage} which means either (a) 'I ask' or (b) 'the question'. Swiss German \textit{fröge, froge, fregu} (a) and or (b) \textit{fraag, froog} were mixed.  (For both examples, see table 1.)

The second case of missing straightforward correspondence is distance between standard German and Swiss German. For one, lexical preferences in Swiss German differ from those in standard German. To express that food is 'tasty' in standard German, the word \textit{lecker} is used. This is also possible in Swiss German, yet the word \textit{fein} is much more common. Another example is that the standard German word \textit{rasch} ('swiftly') is uncommon in Swiss German – synonyms of the word are preferred. Both of this shows in the variety of options our annotators chose for those words (see table 1). Also, the same standard German word may have several dialectal versions in Swiss German. For example there is a short and long version for the standard German word \textit{grossvater}, namely \textit{grospi} and \textit{grossvatter}. 
%or: abonnement – abo. examples too long for table! LS

A second aim was to represent the way Swiss German speaking people write spontaneously. However, as our annotators wrote the spontaneous GSWs mostly while looking at standard German words, our GSWs might be biased towards standard German orthography. Yet, there is potentially also a standard German influence in the way Swiss German is actually written.

We partly revised our dictionary in order to adapt to everyday writing:  We introduced explicit boundary marking into our SAMPAs. We inserted an \_ in the SAMPA where there would usually be a space in writing. An example where people would conventionally add a space are corresponding forms to  standard German preterite forms, for example 'ging'. The Swiss German corresponding past participles – here \textit{isch gange} –  would (most often) be written separately. So entries like b i n k a N @ in table 1 were changed to b i n \_ k a N @.

\section{Conclusion}\label{Conclusion}

In this work we introduced the first Swiss German dictionary. Through its dual nature - both spontaneous written forms in multiple dialects and accompanying phonetic representations - we believe it will become a valuable resource for multiple tasks, including automated speech recognition (ASR). This resource was created using a combination of manual and automated work, in a collaboration between linguists and data scientists that leverages the best of two worlds - domain knowledge and data-driven focus on likely character combinations. 

Through the combination of complementary skills we overcame the difficulty posed by the important variations in written Swiss German and generated a resource that adds value to downstream tasks. We show that the SAMPA to written Swiss German is useful in speech recognition and can replace the previous state of the art. Moreover the written form to SAMPA is promising and has applications in areas like text-to-speech.

We make the dictionary freely available for researchers to expand and use.

\section{Acknowledgements}\label{sec:acknowledgements}

We would like to thank our collaborators Alina Mächler and Raphael Tandler for their valueable contribution.  

\section{Bibliographical References}
\label{main:ref}

\bibliographystyle{lrec}
\bibliography{lrec2020}

\begin{thebibliography}{}

\bibitem[\protect\citename{Bisani and Ney}2008]{Sequir}
Bisani, M. and Ney, H.
\newblock (2008).
\newblock Joint-sequence models for grapheme-to-phoneme conversion.
\newblock {\em Speech Commun.}, 50(5):434--451, May.

\bibitem[\protect\citename{Bohnenberger}1913]{bohnenberger}
Bohnenberger, K.
\newblock (1913).
\newblock {\em Die Mundart der deutschen Walliser im Heimattal und in den
  Aussenorten}.
\newblock Huber \& Co., Frauenfeld.

\bibitem[\protect\citename{Fleischer and Schmid}2006]{FleischerSchmid_2006}
Fleischer, J. and Schmid, S.
\newblock (2006).
\newblock Illustrations of the ipa: Zurich german.
\newblock {\em Journal of the International Phonetic Association},
  36(2):243--253.

\bibitem[\protect\citename{Gasser \bgroup et al.\egroup
  }2010]{gasser_etal_2010}
Gasser, M., Häcki-Buhofer, A., and Hofer, L.
\newblock (2010).
\newblock {\em Neues Baseldeutsch Wörterbuch}.
\newblock Christoph Merian Stiftung, Basel.

\bibitem[\protect\citename{Hollenstein and Aepli}2014]{noah}
Hollenstein, N. and Aepli, N.
\newblock (2014).
\newblock Compilation of a swiss german dialect corpus and its application to
  pos tagging.
\newblock In Marcos Zampieri, et~al., editors, {\em Proceedings of the First
  Workshop on Applying NLP Tools to Similar Languages, Varieties and Dialects
  (VarDial), COLING 2014}, pages 85--94, Dublin, Ireland, august. Association
  for Computational Linguistics and Dublin City University.

\bibitem[\protect\citename{{Honnet} \bgroup et al.\egroup }2017]{honnet-etal17}
{Honnet}, P.-E., {Popescu-Belis}, A., {Musat}, C., and {Baeriswyl}, M.
\newblock (2017).
\newblock {Machine Translation of Low-Resource Spoken Dialects: Strategies for
  Normalizing Swiss German}.
\newblock {\em ArXiv e-prints}, 1710.11035, October.

\bibitem[\protect\citename{Hotzenk{\"o}cherle \bgroup et al.\egroup }1962
  1997]{sds}
Rudolf Hotzenk{\"o}cherle, et~al., editors.
\newblock (1962--1997).
\newblock {\em Sprachatlas der deutschen Schweiz}.
\newblock Francke, Bern.

\bibitem[\protect\citename{Hug and Weibel}2003]{Hug_weibel_2003}
Hug, A. and Weibel, V., (2003).
\newblock {\em Lautliche Verhältnisse}, pages 30--33.

\bibitem[\protect\citename{Kolly and Leemann}2015]{dialaekt_app}
Kolly, M.-J. and Leemann, A., (2015).
\newblock {\em Dialäkt App: communicating dialectology to the public –
  crowdsourcing dialects from the public}, pages 271--285.
\newblock Peter Lang, Bern.

\bibitem[\protect\citename{Kolly \bgroup et al.\egroup }2014]{voiceapp}
Kolly, M.-J., Leemann, A., Dellwo, V., Goldman, J.-P., Hove, I., and Ibrahim,
  A.
\newblock (2014).
\newblock ‘swiss voice app’: A smartphone application for crowdsourcing
  swiss german dialect data.
\newblock In {\em Digital Humanities}, pages 231--233, Lausanne, july.

\bibitem[\protect\citename{Lusetti \bgroup et al.\egroup }2018]{lusettietal_18}
Lusetti, M., Ruzsics, T., Göhring, A., Samardžić, T., and Stark, E.
\newblock (2018).
\newblock Encoder-decoder methods for text normalization.
\newblock In {\em Proceedings of the Fifth Workshop on NLP for Similar
  Languages, Varieties and Dialects (VarDial 2018)}, pages 18--28, Santa Fe,
  New Mexico, USA, August. University of Zurich.

\bibitem[\protect\citename{Marti}1985]{marti_1985}
Marti, W.
\newblock (1985).
\newblock {\em Berndeutsch-Grammatik : Für die heutige Mundart zwischen Thun
  und Jura}.
\newblock A. Francke, Bern.

\bibitem[\protect\citename{Niederberger}2007]{niederberger_2007}
Niederberger, E.
\newblock (2007).
\newblock {\em Nidwaldner Mundart: Wörterbuch von Ernst Niederberger}.
\newblock Edition Odermatt, Dallenwil, 3rd edition.

\bibitem[\protect\citename{Osterwalder-Brändle}2017]{osterwalder-braendle_2017}
Osterwalder-Brändle, S.
\newblock (2017).
\newblock {\em Hopp Sanggale! St. Galler Mundartwörterbuch: Wörter,
  Redensarten, Geschichten}.
\newblock Cavelti AG, Gossau.

\bibitem[\protect\citename{Ruzsics \bgroup et al.\egroup }2019]{ruzsicsetal_19}
Ruzsics, T., Lusetti, M., Göhring, A., Samardžić, T., and Stark, E.
\newblock (2019).
\newblock Neural text normalization with adapted decoding and pos features.
\newblock {\em Natural Language Engineering}, 25(5):585--605.

\bibitem[\protect\citename{Samard{\v z}i{\' c} \bgroup et al.\egroup
  }2015]{Samardzicetal2015}
Samard{\v z}i{\' c}, T., Scherrer, Y., and Glaser, E.
\newblock (2015).
\newblock Normalising orthographic and dialectal variants for the automatic
  processing of {Swiss German}.
\newblock In {\em Proceedings of The 4th Biennial Workshop on Less-Resourced
  Languages}, pages 294--298. ELRA.

\bibitem[\protect\citename{Samard{\v z}i{\'c} \bgroup et al.\egroup
  }2016]{archimob_16}
Samard{\v z}i{\'c}, T., Scherrer, Y., and Glaser, E.
\newblock (2016).
\newblock Archi{M}ob - a corpus of spoken {S}wiss {G}erman.
\newblock In Nicoletta Calzolari~(Conference Chair), et~al., editors, {\em
  Proceedings of the Tenth International Conference on Language Resources and
  Evaluation (LREC 2016)}, Paris, France, may. European Language Resources
  Association (ELRA).

\bibitem[\protect\citename{Samardžić \bgroup et al.\egroup
  }2015]{archimob_15}
Samardžić, T., Scherrer, Y., and Glaser, E.
\newblock (2015).
\newblock Normalising orthographic and dialectal variants for the automatic
  processing of swiss german.
\newblock In {\em Proceedings of the 7th Language and Technology Conference:
  Human Language Technologies as a Challenge for Computer Science and
  Linguistics}, pages 294--298, Poznan, Poland, November.

\bibitem[\protect\citename{Scherrer and
  Ljube\v{s}i\'{c}}2016]{scherrer16-automatic}
Scherrer, Y. and Ljube\v{s}i\'{c}, N.
\newblock (2016).
\newblock {Automatic normalisation of the Swiss German ArchiMob corpus using
  character-level machine translation}.
\newblock In {\em Proceedings of the 13th Conference on Natural Language
  Processing (KONVENS 2016)}, pages 248--255.

\bibitem[\protect\citename{Scherrer and Stöckle}2016]{scherrer_stoeckle16}
Scherrer, Y. and Stöckle, P.
\newblock (2016).
\newblock A quantitative approach to swiss german – dialectometric analyses
  and comparisons of linguistic levels.
\newblock {\em Dialectologia et Geolinguistica}, 24(1):92--125.

\bibitem[\protect\citename{Scherrer \bgroup et al.\egroup }2019]{archimob_19}
Scherrer, Y., Samardžić, T., and Glaser, E.
\newblock (2019).
\newblock Digitising swiss german -- how to process and study a polycentric
  spoken language.
\newblock {\em Language Resources and Evaluation}, pages 1--35.

\bibitem[\protect\citename{Scherrer}2012]{scherrer_12}
Scherrer, Y.
\newblock (2012).
\newblock Generating swiss german sentences from standard german: A
  multi-dialectal approach. {D}octoral {T}hesis.
\newblock \url{https://doi.org/10.13097/archive-ouverte/unige:26361}.

\bibitem[\protect\citename{Stark \bgroup et al.\egroup }2009  2015]{sms_corpus}
Stark, E., Ueberwasser, S., and Ruef, B.
\newblock (2009--2015).
\newblock Swiss {SMS} corpus, {U}niversity of {Z}urich.
\newblock \url{https://sms.linguistik.uzh.ch}.

\bibitem[\protect\citename{Stark \bgroup et al.\egroup }2014]{wus_14}
Stark, E., Ueberwasser, S., and Göhring, A.
\newblock (2014).
\newblock Corpus "what's up switzerland". technical report.
\newblock Switzerland. University of Zurich.

\bibitem[\protect\citename{Stark}2014  ]{wus_corpus}
Stark, Elisabeth;~Ueberwasser, S. G.~A.
\newblock (2014--).
\newblock Korpus 'what’s up, switzerland?'.
\newblock \url{https://www.whatsup-switzerland.ch}.

\bibitem[\protect\citename{Staub \bgroup et al.\egroup }1881  ]{idiotikon}
Friedrich Staub, et~al., editors.
\newblock (1881--).
\newblock {\em Schweizerisches Idiotikon: W{\"o}rterbuch der schweizerdeutschen
  Sprache}.
\newblock Huber, Frauenfeld 1881--2012 / Schwabe, Basel 2015ff.

\bibitem[\protect\citename{Suter}1992]{suter_1992}
Suter, R.
\newblock (1992).
\newblock {\em Baseldeutsch-Grammatik}.
\newblock C. Merian, Basel.

\bibitem[\protect\citename{Vaswani \bgroup et al.\egroup
  }2017]{vaswani2017attention}
Vaswani, A., Shazeer, N., Parmar, N., Uszkoreit, J., Jones, L., Gomez, A.~N.,
  Kaiser, {\L}., and Polosukhin, I.
\newblock (2017).
\newblock Attention is all you need.
\newblock In I.~Guyon, et~al., editors, {\em Advances in Neural Information
  Processing Systems 30}, pages 5998--6008.

\end{thebibliography}

\end{document}